# `destroR`: A Benchmark and Adversarial-Training Defense for Bangla Transfer Models under Meaning-Preserving Attacks

Saadat Rafid Ahmed, Rubayet Shareen, Radoan Sharkar, Nazia Hossain, Mansur Mahi
*Dept. of Computer Science and Engineering, BRAC University*
*Email: {saadat.rafid.ahmed, rubayet.shareen, radoan.sharkar, nazia.hossain1, mansur.mahi}@g.bracu.ac.bd*

Dr. Farig Yousuf Sadeque
*Assistant Professor, Dept. of Computer Science and Engineering, BRAC University*
*Email: farig.sadeque@bracu.ac.bd*

***Abstract*—Transformer-based transfer models now dominate Bangla sentiment classification, yet their adversarial robustness remains largely unexamined, and no prior study pairs a Bangla attack suite with a defense that measurably recovers robustness. We address this gap with `destroR`, a unified pipeline for evaluating and hardening Bangla text classifiers. First, we introduce three meaning-preserving Bangla attack recipes—a paraphrase attack, a back-translation attack, and a one-hot word-swap attack—that perturb inputs while regenerating fluent, semantically faithful sentences, inducing model prediction perplexity rather than input noise. Second, we construct a robustness benchmark that evaluates five transfer models (BanglaBERT, BanglishBERT, XLM-RoBERTa, MuRIL, and IndicBERTv2) across four datasets against five attacks, placing our recipes against two strong word-substitution baselines, TextFooler and BAE, under an identical protocol. Third, we harden every model through adversarial training and report a full robustness matrix. Our analysis yields three findings: word-substitution baselines are more potent than semantically constrained recipes (BAE reaches a 54.2% attack success rate); adversarial training on the union of all attack families lowers residual attack success for every attack; and, contrary to expectation, the Indic-multilingual MuRIL backbone is markedly more robust than the Bangla-dedicated models. All models, adversarial data, and code are released for full reproducibility.**

## 1. Introduction

Neural networks and transformer architectures have become the default tools for natural language processing, and Bangla—a language spoken by more than 230 million people—has benefited from a recent wave of pretrained transfer models that reach strong accuracy on tasks such as sentiment classification. High benchmark accuracy, however, is not the same as robustness. A growing body of work shows that text classifiers can be reliably fooled by adversarial examples: minimally altered inputs that preserve the meaning a human reads but flip the label a model predicts. For deployed systems that moderate content, gauge public opinion, or triage user feedback, this brittleness is a genuine reliability and security concern.

Most adversarial NLP research targets English. Bangla, despite its speaker base, is a low-resource language for this purpose: it lacks the lexical databases, paraphrase corpora, and evaluation tooling that English attacks take for granted, and its rich morphology, frequent code-mixing, and transliteration make attack design non-trivial. As a result, the robustness of Bangla transfer models has not been systematically characterised, and no prior study couples a Bangla attack suite with a defense that is shown to recover robustness.

### 1.1. Problem Statement

Creating adversarial examples is well established in computer vision but comparatively new in NLP, where the discrete, sequential nature of text makes imperceptible perturbation difficult: any edit is visible, and preserving meaning while changing the prediction requires linguistic care. For Bangla the difficulty is compounded by the scarcity of supporting resources. The central question we investigate is:

***How robust are Bangla transfer models to adversarial attacks, and can adversarial training recover that robustness?***

### 1.2. Contributions

To answer this question we build `destroR`, a unified pipeline for evaluating and hardening Bangla sentiment classifiers. Our contributions are:

1) **Three meaning-preserving Bangla attack recipes** — a paraphrase attack, a back-translation attack, and a one-hot word-swap attack — that regenerate fluent, semantically faithful sentences rather than injecting surface noise.
2) **A robustness benchmark** evaluating five transfer models across four datasets against five attacks, situating our recipes against two strong word-substitution baselines (TextFooler and BAE) under an identical protocol.
3) **An adversarial-training defense** with a full robustness matrix, quantifying how much each training regime hardens each model against each attack.
4) **A reproducible release** of all victim models, generated adversarial data, and code, with fixed seeds and logged configurations.

We deliberately frame our recipes as *meaning-preserving*: they aim to keep a sentence's human-perceived semantics intact while driving the model into a state of prediction perplexity—collapsing its confidence or flipping its label. This separates input fidelity from model perplexity, a distinction we make precise in the threat model of Section 3.

## 2. Related Work

### 2.1. Adversarial Attacks on Text Classifiers

Textual adversarial attacks are commonly organised by their level of access to the victim and by the granularity of their perturbation. White-box attacks exploit model internals: HotFlip [1] uses the gradient of one-hot input vectors to select character- and word-level flips that maximise loss, and universal adversarial triggers [2] search for input-agnostic token sequences that induce systematic errors. Black-box attacks assume only query access. TextBugger [3] generates character- and word-level perturbations effective in both settings, and two black-box word-substitution methods have become standard baselines: TextFooler [4], which ranks words by importance and substitutes synonyms constrained by embedding and sentence-encoder similarity, and BAE [5], which uses a masked language model to generate context-aware replacements. Interpretability tools underpin many of these methods—feature-attribution explanations [6], influence functions [7], and toolkits such as AllenNLP Interpret [8]—by identifying which tokens most influence a prediction and are therefore worth perturbing. Our one-hot word-swap recipe follows this importance-ranking-then-substitution paradigm, while our paraphrase and back-translation recipes operate at the sentence level, regenerating whole sentences rather than editing individual tokens.

### 2.2. Defenses and Adversarial Training

The most widely used defense is adversarial training: augmenting the training set with generated adversaries so the model learns to resist them. The TextAttack framework [9] standardises both attack generation and adversarial-training pipelines and provides the conceptual template for our evaluation harness. Behavioural testing such as CheckList [10] complements accuracy-based evaluation by probing specific model capabilities. Despite the prominence of adversarial training in English NLP, defense results are almost entirely absent from prior Bangla adversarial work; a central contribution of this paper is to report them.

### 2.3. Low-Resource and Multilingual Adversarial NLP

Attacks designed for English rely on resources—WordNet, large paraphrase corpora, strong sentence encoders—that are scarce or absent in most languages. Multilingual sentence encoders such as LaBSE [11] partially close this gap by providing cross-lingual similarity estimates that are markedly stronger than earlier universal encoders on non-English text, which is why we adopt LaBSE for the semantic-fidelity constraint. Cross-lingual attacks on machine translation, such as TransFool [12], further show that the translation process itself can be turned into an adversarial mechanism—an observation that directly motivates our back-translation recipe.

### 2.4. NLP for Bangla and Indic Languages

Recent pretrained models have substantially improved Bangla and Indic NLP. BanglaBERT [13] and its code-mixed variant BanglishBERT are ELECTRA-style models pretrained on large Bangla corpora; XLM-RoBERTa [14] provides broad multilingual coverage; and MuRIL [15] and IndicBERTv2 [16] target Indic languages specifically. On the generation side, BanglaT5 and the BanglaParaphrase dataset [17] supply the paraphrasing and translation backbones our sentence-level attacks depend on, while efforts such as BanglaNet [18] illustrate the still-limited state of Bangla lexical resources. We evaluate the first four model families above as victims and use the BanglaT5 backbones to construct our attacks.

## 3. Preliminaries and Threat Model

We consider sentiment classification: a victim model $f$ maps a Bangla input $x$ to a label $f(x) \in \{\text{positive}, \text{negative}, \text{neutral}\}$ with confidence $s(x)$. An adversary transforms $x$ into $x'$ with the goal of changing the prediction while preserving the meaning a human reader assigns to the sentence. We formalise the setting along four axes.

Table 1: Positioning of `destroR` against representative textual adversarial-attack methods. `destroR` is distinguished by its Bangla focus, its combination of word- and sentence-level perturbations, and the fact that it reports a defense.

| **Method** | **Granularity** | **Access** | **Language** | **Constraint** | **Defense shown** |
|---|---|---|---|---|---|
| TextFooler [4] | Word | Black | EN | USE similarity | No |
| BAE [5] | Word | Black | EN | BERT MLM | No |
| TextBugger [3] | Char/Word | Both | EN | Edit distance | No |
| HotFlip [1] | Char/Word | White | EN | Gradient | No |
| TransFool [12] | Word | White | EN–FR/DE | Similarity | No |
| `destroR` (ours) | **Word + Sentence** | **Black/Gray** | **Bangla** | **LaBSE + BERT MLM** | **Yes** |

Attacker's goal. The attack is *untargeted*: the adversary seeks any label change, $f(x') \neq f(x)$, on inputs the model originally classifies correctly. We measure this with the Attack Success Rate (ASR), the fraction of originally correct predictions an attack flips. Where a label is not changed, we additionally track the drop in prediction confidence, since a confident model driven to near-chance confidence is a partial success—a state we refer to as model *perplexity*.

Attacker's knowledge. Our paraphrase and back-translation recipes are *black-box*: they require only the model's output label and confidence. The one-hot word-swap recipe is *gray-box*: it queries the model to rank token importance via leave-one-out and then fills masked positions with a language model, but does not access gradients or weights. The two baselines, TextFooler and BAE, are likewise black-box query-based attacks.

Attacker's capability. We distinguish input fidelity from model perplexity. A valid adversary must preserve semantics: we constrain candidates with LaBSE cosine similarity $\mathrm{sim}_{\mathrm{LaBSE}}(x, x') \geq \tau$ (with $\tau = 0.7$) and, for the word-level recipe, cap the perturbation at 20% of tokens. Formally, the one-hot recipe seeks

$$x_{\mathrm{adv}} = \arg\max_{x' \in \mathcal{N}(x)} \mathcal{L}\big(f(x'), y\big) \quad \text{s.t.} \quad \mathrm{sim}_{\mathrm{LaBSE}}(x, x') \geq \tau,$$

where $\mathcal{N}(x)$ is the masked-language-model neighbourhood of $x$ and $\mathcal{L}$ is the classification loss. The sentence-level recipes replace $\mathcal{N}(x)$ with the set of fluent paraphrases or back-translations of $x$.

Defender's capability. The defender has access to the attack at training time and may augment the training set with generated adversaries (adversarial training). We evaluate three regimes—augmentation with paraphrase adversaries, with back-translation adversaries, and with the union of all attack families—and measure the residual ASR of every attack against the hardened model.

## 4. Data Analysis

In the course of our research, we have identified four datasets serving distinct purposes. One of these datasets is directly associated with the fine-tuned models that form the basis of our research. The remaining three datasets have been curated to introduce additional variations in the dynamics of the data, allowing for a comprehensive examination of the robustness of the models. The names of the selected datasets are as follows: blp23[1], youtube_sentiment, CogniSenti, and BASA_cricket[2]. We chose blp23 and youtube_sentiment for understanding the performance difference between internal and external datasets, while CogniSenti and BASA_cricket was chosen for understanding the difference of performance of the attack methodology based on the average length of a dataset. The detials of the dataset can be found in the table 2.

## 5. Attack Methodology

### 5.1. Bangla Paraphrase Attack

To come up with a sentence-level adversarial methodology, we employed the capabilities of a paraphrasing tool. We came across multiple possible methods of creating a paraphrase augmentation. However, we chose the csebuetnlp/banglat5_banglaparaphrase, a finetuned version of their own BanglaT5 model for providing diverse and context-aware paraphrases of Bangla sentences [17]. The high PINC score (0.65, 0.76, 0.80) ensured us of diverse adversaries, and the BERTScore(lower 0.91 - 0.93, upper 0.98) provided trustworthiness in preserving the coherence with the original text while being aware of the context in bigram to quad-gram awareness [17]. When paraphrased, the structure of a sentence usually changes along with the lexical state. By generating paraphrased adversaries, we expect to see some mispredictions or a decreased performance of LLMs by attacking certain models and understanding their capability of handling different sentence structures without changing the semantics. Thus, we created our Bangla Paraphrase Attack (Algorithm 1).

### 5.2. Bangla Back Translation

Back Translation has always been one of the popular tasks in NLP. That is why we can come across many resources on Bangla translation. Translation creates a semantic bridge between two languages; however, it

1. https://github.com/blp-workshop/blp_task2
2. https://github.com/banglanlp/bangla-sentiment-classification

| Dataset | Source | Negative | Neutral | Positive | Total |
|---|---|---|---|---|---|
| youtube_sentiment | YouTube Comments | 865 | 539 | 553 | 1957 |
| BASA_Cricket | Social Media and Newspaper Articles | 1515 | 194 | 376 | 2085 |
| CogniSenti | Social Media | 919 | 2633 | 1047 | 4599 |
| blp23 | Social Media | 15767 | 7135 | 12364 | 35266 |

Table 2: All Dataset Description

**Algorithm 1:** Bangla Paraphrase Attack

```
Input: model, dataset, labelMap=None,
       csvFileName=None,
       generate_report=False
Output: dataset_attacked
1  initialize_columns(dataset)
2  for data point to dataset do
3      init_pred, init_score :=
         model(data_point['text'], labelMap)
4      if data_point['label'] ≠
         data_point['initPred'] then
5          data_point['atckSuccess'] :=
             'Misprediction'
6          continue
7      augmented_text :=
         paraphrase_augment(data_point['text'])
8      atck_pred, atck_score :=
         model(augmented_text, labelMap)
9      data_point['atckPred'] := atck_pred
10     data_point['atckScore'] := atck_score
11     data_point['atckSent'] := augmented_text
12     if data_point['atckPred'] ==
         data_point['initPred'] then
13         data_point['atckSuccess'] := 'Fail'
14     else
15         data_point['atckSuccess'] := 'Pass'
16 if generate_report then
17     print_report(initial_model_performance(dataset))
18     print_report(performance_after_attack(dataset))
19 save_to_csv(dataset, csvFileName)
20 return filtered_dataset(dataset)
```

often fails to maintain the same sentence structure due to the difference in alignment property of the machine translation models. We used this phenomenon to our advantage and got inspired by the adversarial attack, TransFool [12]. We tried to find a tool to provide us with the capabilities to achieve this methodology. So, the pipeline should look like this(5.2),

**Theoritical Idea of Bangla Back Translation**

Init $\text{Lang}_1 \Rightarrow$ Intermediate Lang $\Rightarrow$ Init $\text{Lang}_2$

Here, Init $\text{Lang}_1$ and Init $\text{Lang}_2$ are in the same language but theoretically different in Structure/Lexicon

We chose English to be our intermediate language, as it has the most resources on Bangla to English translation. For this reason, we used 'csebuetnlp/banglat5_nmt_en_bn' and 'csebuetnlpbanglat5_nmt_bn_en'. This pipeline of two way translation helped us create the Bangla Back Translation Attack (Algorithm 2) which should provide us with semantically close adversaries with different structures and lexicons.

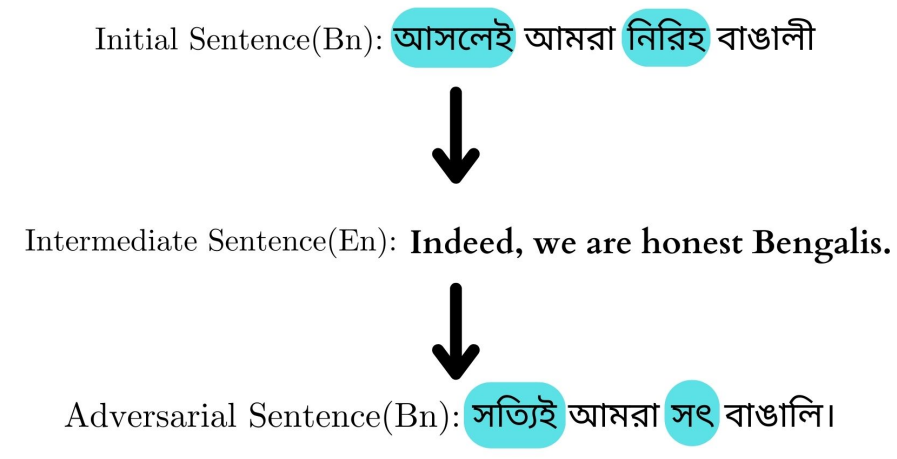

Figure 1: Back-translation illustrative idea

### 5.3. One-Hot Word Swap Attack

Regarding sentence classification, the classifier may disproportionately assign weight to a specific token, influenced by its frequent repetition in the Language Model (LLM) training dataset. This phenomenon introduces bias in LLMs, favoring the emphasized token and potentially affecting the model's overall performance and generalization. Our hypothesis for this attack development is to harness this bias and perturb it with another augmented text to obscure the model we are attacking. The masked language modeling capability of Bidirectional Encoder Representations from Transformers (BERT) provides us with the weapon to attack these LLMs. We identify these important aka. Biased tokens and replace them with the unmasked tokens we get from the BERT model. We experimented with multiple BERTs but finalized with

**Algorithm 2:** Bangla Back Translation Attack

```
Input: model, dataset, labelMap=None,
       csvFileName=None,
       generate_report=False
Output: [status, attacks]
1  initialize_columns(dataset)
   /* Attacking                                        */
2  for data point to dataset do
3      init_pred, init_score :=
        model(data_point['text'], labelMap)
4      if data_point['label'] ≠
        data_point['initPred'] then
5          data_point['atckSuccess'] :=
            'Misprediction'
6          continue
7      augmented_text :=
        raw_back_translation_augment(data_point['text'])
8      atck_pred, atck_score :=
        model(augmented_text, labelMap)
9      data_point['atckPred'] := atck_pred
10     data_point['atckScore'] := atck_score
11     data_point['atckSent'] := augmented_text
12     if data_point['atckPred'] ==
        data_point['initPred'] then
13         data_point['atckSuccess'] := 'Fail'
14     else
15         data_point['atckSuccess'] := 'Pass'
   /* Generating Report                                */
16 if generate_report then
17     print_report(initial_model_performance(dataset))
18     print_report(performance_after_attack(dataset))
19 save_to_csv(dataset, csvFileName)
20 return filtered_dataset(dataset)
```

the xlm-roberta-base, pre-trained on 2.5TB of filtered CommonCrawl data containing 100 languages. It contains 525 Bangla tokens and 77 Bangla Romanized tokens, making a size of 8.4 GiB and 0.5 GiB [14]. As this dataset contains multiple languages, it provides us with a large domain of tokens and helps the attacking method to create much more profound unmasked tokens. Thus, we created the One-Hot Word Swap Attack (Algorithm 3 & 4) that contributed to the core of this word-level attack methodology.

## Victim Models and Baseline Attacks

To characterise robustness across model families rather than a single classifier, we finetune five transfer models on each dataset with identical hyperpa-

**Algorithm 3:** One-Hot Word Swap for a Single Sentence

```
Input: model, tok, init_pred, init_score,
       label_map=None
Output: dataset_attacked
   /* Prioritizing each candidate token                */
1  for i ← 0 to length(tok) - 1 do
2      new := join_tokens_except_i(tok, i)
3      p, s := model(new, label_map)
4      if p ≠ init_pred then
5          score.append([i,
            replace_i_with_mask(tok, i), p, ∞])
           // Most significant perturbation candidate
6      else
7          if s < init_score then
8              score.append([i,
                replace_i_with_mask(tok, i), p, s,
                round((init_score - s) × 100)])
   /* Sort the score list                              */
9  score := sort_score_by_last_element_descend-
    ing(score)
   /* Initialize empty lists                           */
10 attacks := empty list
11 status := 'Fail'
12 unmasker := create_pipeline('fill-mask',
    model='xlm-roberta-base')
   /* Iterate over scores                              */
13 for i ← 0 to length(score) - 1 do
14     u := unmasker(score[i][1])
       /* Iterate over unmasked sequences              */
15     for a in u do
16         p, s := model(a['sequence'], label_map)
17         if p ≠ init_pred then
18             status := 'Pass'
19             attacks.append((a['sequence'], p, s,
                'Pass'))
               // Flip of Initial Prediction = Passed
                  Attack
20         else
21             attacks.append((a['sequence'], p, s,
                'Fail'))
   /* Sort and select the top 10 successful attacks
      */
22 attacks := sort_successful_attack(attacks)
23 attacks :=
    take_top_10_successful_attacks(attacks)
24 return [status, attacks]
```

**Algorithm 4:** One-Hot Word Swap Attack

```
Input: model, dataset, max_pass,
       labelMap=None, csvFileName=None,
       generate_report=False
Output: dataset_attacked
                  // holding all the performed attacks
1  tokenizer := BasicTokenizer()
2  dataset['tok'] := dataset['text'].apply(tokenizer)
   /* Initializing the DataFrame                          */
3  df := create_empty_dataframe(columns=['id',
    'text', 'label', 'initPred', 'initScore', 'atckPred',
    'atckScore', 'atckSuccess', 'atckSent'])
   /* Attacking                                           */
4  number_success := 0
5  for row to range(len(dataset)) do
6      iP, iS := model(dataset.text[row], labelMap)
7      if iP ≠ dataset.label[row] then
8          df.loc[len(df)] := 'id': dataset['id'][row],
            'text': dataset['text'][row], 'label':
            dataset['label'][row], 'initPred': iP,
            'initScore': iS, 'atckPred': '-',
            'atckScore': '-', 'atckSuccess':
            'Misprediction', 'atckSent': '-'
9          continue
10     else
11         status, attacks := atck_sentence(model,
            dataset['tok'][row], iP, iS,
            label_map=labelMap)
12         if status == 'Pass' then
13             number_success := number_success
                + 1
14             print(Total number of Successful
                Augmentation of Datapoints:
                number_success)
15         if number_success == max_pass then
16             break
17
18         for a in attacks do
19             df.loc[len(df)] := 'id':
                dataset['id'][row], 'text':
                dataset['text'][row], 'label':
                dataset['label'][row], 'initPred': iP,
                'initScore': iS, 'atckPred': a[1],
                'atckScore': a[2], 'atckSuccess': a[3],
                'atckSent': a[0]
   /* Generating Result                                   */
20 if csvFileName ≠ None then
21     dataset.to_csv(csvFileName)
22 if generate_report then
23     Generate_Report()
24 return dataset
```

Original Sentence: মাল যথার্থ ঠিক কথাই বলেছেন (53% Pos)
Sentence Masked: মাল যথার্থ ঠিক {mask} বলেছেন
MLM
P1: করে
P2: মত
S1(P1): মাল যথার্থ ঠিক করে বলেছেন (52% Neu)
S2(P2): মাল যথার্থ ঠিক মত বলেছেন (47% Neu)

Figure 2: One-Hot BERT Perturb illustrative idea

rameters and treat them as victims: BanglaBERT [13] and its code-mixed variant BanglishBERT, both ELECTRA-style models pretrained on Bangla; XLM-RoBERTa [14], a broadly multilingual baseline; MuRIL [15], an Indic-specific model; and IndicBERTv2 [16]. These span Bangla-dedicated, Indic-multilingual, and general-multilingual pretraining, letting us test whether language-specific pretraining confers robustness. To place our recipes in context, we additionally evaluate two strong black-box word-substitution baselines from the English literature under an identical protocol: TextFooler [4], which substitutes importance-ranked words with similarity-constrained synonyms, and BAE [5], which fills masked positions with a BERT masked language model. All victims are built with the Hugging Face classification pipeline.

## 6. Result Analysis

### 6.1. Quantitative Analysis of Attacks

We begin with a detailed case study on a single victim—a BanglaBERT model finetuned on the blp23 dataset for ternary sentiment classification—before presenting the full multi-model benchmark in Section 6.3. This case study lets us examine the mechanism of each attack at the level of individual predictions and confidence scores. We attacked the model with both an external dataset (the YouTube-sentiment set, which the model was not trained on) and an internal dataset (the set the model was trained on), so that the attack's effect could be judged against both the domains the training data covered and those it did not. In every case we observed a measurable decrease in prediction confidence, and we found that the internal dataset drove the model into a greater state of perplexity than the external one.
The details of this attack ka05ar/banglabert-sentiment[3] can be found in Table 4. In the case of One-Hot BERT Perturb, we could observe that it creates the most amount of adversarial examples; that is, for each attackable datapoint, at most ten adversaries were created, and reduced the performance of the model(F1 Score macro) by 37% on the External dataset and 40% on the

3. https://huggingface.co/ka05ar/banglabert-sentiment

| Attack Methodology | Dataset | Attacks | | Initial Score | Attack Score | DIfference |
|---|---|---|---|---|---|---|
| | | Datapoints | % Successful | F1 Score Macro | F1 Score Macro | F1 Score |
| **Bangla Paraphrase Attack** | **External** | 304 | 12.83 | 100% | 82% | 18% |
| | **Internal** | 344 | 13.37 | 100% | 78% | 22% |
| **Back Translation** | **External** | 304 | 17.76 | 100% | 78% | 22% |
| | **Internal** | 72 | 13.89 | 100% | 77% | 23% |
| **One-Hot BERT Perterb** | **External** | 160 | 25.63 | 100% | 63% | 37% |
| | **Internal** | 190 | 30.53 | 100% | 60% | 40% |

Table 3: Attack on *k05ar/banglabert-sentiment*

internal dataset. Similarly, for the Bangla Paraphrase attack, we observed 18%(External Dataset) and 22%(Internal Dataset), and In Back translation, we observed a 22%(External) and 23%(Internal) reduction in the F1 Score. However, the relation between attachable data points and the adversarial examples generated is 1:1. From this analysis; we can consider that the world-level One-Hot BERT perturb attack is the most sophisticated attack as the mean reduction of performance is the greatest in this case among all the attacking methodology. This performance reduction is much expected, as the One-Hot Bert Perturb may sometimes be delusional and not always create a sound and non-inverting adversary, and the other sentence-level attacks attack the model of lexical and structural modification.
Three cases can emerge when attacking a model based on a data point. The attack may cause three cases for each data point: Pass, Fail, or Misprediction. For the passed attacks, the models prediction is inverted. However, the mean confidence level of the models prediction is around 50.54%, which can be questionable. This can be considered a state of perplexity or confusion for the model. Here, the actual goal of the attacks is achieved. Furthermore, we may not see a prediction label inversion when we look into the failed attacks. However, we still observe a reduction in confidence level, which could also be a subtle achievement of our attack algorithms. We observed that the mean performance of the model was reduced from 77.29% to 74.71%. When a model already makes a misprediction for a particular data point, we do not attack it. It can already be considered a passed attack.

## 6.2. Illustrative Analysis of Attacks

To illustrate the examples of different types we have constructed a table consisting of eight attributes -

- Original Sentence: s1
- Initial Prediction: $\{\text{pos}, \text{neg}, \text{neu}\}$
- Initial Score: x
- Augmented Sentence: augmented (s1)
- Prediction on Augmentation: $\{\text{pos}, \text{neg}, \text{neu}\}$
- Score after Augmentation: y
- Delta ($\delta$): $(\text{x} - \text{y}) \times 100$
- Verdict: Pass/Fail

Here, the delta ($\delta$) value plays a significant role. As we have seen, not all of our attacks were successful, and we wanted to know how far our attack has failed. To measure that, the delta value is introduced.
The delta, $\delta$, is computed as follows:

$$delta, \delta = (initialScore - augmentedScore) \times 100$$

Point to be noted: the delta value is only applicable to failed attacks, the attack that was unable to change the initial label prediction of the model. Thus, for successful attacks, the delta value is N/A.
The change in delta value is associated with the change in the confidence score of the model's prediction. If the delta value is negative, it means the model gains confidence after the attack is performed, and vice versa.

### 6.2.1. Bangla Paraphrase Attack.

Here are some examples with all the attributes for Bangla Paraphrase Attack-
From Table 5, Example 1 illustrates that the initial sentence is predicted to be negative. However, after data augmentation, the model mis-predicts the class label as positive. In this case, the attack successfully confused the model because the sentence before and after augmentation conveys the same meaning, and both should have been labeled as negative sentiments. The two changes added to the augmented sentence are the word "একজন" and the transliteration "Pagol" in the original text, which do not change the semantics but rather make the structure more profound and formal. A similar scenario can be observed in Example 2. However, in Example 3, even though the attack was successful, we can see that the augmentation altered the meaning of the original sentences. In Examples 4 and 5, the attack failed; however, a drastic change in the confidence score of the prediction is noticeable.

### 6.2.2. Bangla Back Translation Attack.

Here are some examples with all the attributes for Bangla Back Translation Attack-
In Table 6 Examples 1 and 2, we were able to fool the model with our augmentation. In both cases, it can be observed that the lexical structure of the sentence changes while keeping the semantic meaning of the original sentence intact. In Example 3, transliteration takes place in the augmented sentence, and this alone changes the prediction of the model.
In Examples 4 and 5, even though the augmentations were successful, the attack did not manage to fool the model. In some of the augmentations, it was observed that the original sentence was retranslated to exactly the same sentence. This can be seen in Example 6. In such cases, we can conclude that the augmentation was unsuccessful.

| Attack Number and Attack | | 1.1 | 1.2 | 1.3 | 1.4 | 1.5 | 1.6 | Previous confidence of failed attacks | After Attack Average |
|---|---|---|---|---|---|---|---|---|---|
| Passed Attack Prediction Confidence | Mean | 51.6 | 50.75 | 52.49 | 52.89 | 50.36 | 45.16 | | 50.54 |
| | Median | 48.71 | 45.65 | 51.27 | 49.39 | 47.38 | 43.3 | | 47.62 |
| | Mode | 44.34 | 34.58 | 43.68 | 49.86 | 50.59 | 42.87 | | 44.32 |
| Failed Attack Prediction Confidence | Mean | 73.56 | 71.43 | 82.68 | 78.43 | 71.99 | 70.19 | **77.29** | **74.71** |
| | Median | 76.75 | 73.92 | 86.65 | 84.03 | 77.51 | 76.69 | **82.60** | **79.26** |
| | Mode | 54.58 | 65.4 | 76.65 | 85.91 | 52.89 | 79.3 | **74.49** | **69.12** |
| Misprediction Score | Mean | 58.85 | 59.74 | 63.2 | 57.63 | 58.85 | 57.81 | 59.65 | |
| | Median | 53.59 | 56.23 | 55.6 | 54.51 | 53.59 | 57 | 54.70 | |
| | Mean | 83.01 | 44.16 | 83.01 | 44.16 | 83.01 | 44.16 | 67.47 | |

Table 4: Before vs After Attack Confidence *ka05ar/banglabert-sentiment*

| | | Sentences | Prediction | Score | % $\delta$ \| Result |
|---|---|---|---|---|---|
| 1 | Original | শাহীন আখাউডা সৌদি প্রবাসী pagol | Negative | 0.90 | N/A \| P |
| | Augmented | শাহিন আখাউডা একজন সৌদি প্রবাসী পাগল। | Neutral | 0.46 | |
| 2 | Original | এই বটম জেস্চার মেবি প্রথম আনে ব্ল্যাকবেরি! | Neutral | 0.46 | N/A \| P |
| | Augmented | এই বটম জেসচারই প্রথম ব্ল্যাকবেরি নিয়ে আসে! | Positive | 0.66 | |
| 3 | Original | হ গেছে পুরাই | Negative | 0.41 | N/A \| P |
| | Augmented | সে পুরোপুরি ঠিক আছে। | Positive | 0.60 | |
| 4 | Original | আমার দেশ বাংলাদেশ । আমি গর্বিত | Positive | 0.38 | -49.22 \| F |
| | Augmented | আমার দেশ বাংলাদেশ, আমি গর্বিত। | Positive | 0.88 | |
| 5 | Original | রাশিয়ায় বড় জয়ের পথে পুতিনের দল | Positive | 0.89 | 51.51 \| F |
| | Augmented | পুতিনের দল রাশিয়ার বিশাল বিজয়ের পথে। | Positive | 0.37 | |
| 6 | Original | সালাম স্যার | Positive | 0.82 | 18.79 \| F |
| | Augmented | সালাম স্যার। | Positive | 0.64 | |

Table 5: Bangla Paraphrase Attack Examples

| | | Sentences | Prediction | Score | % $\delta$\| Result |
|---|---|---|---|---|---|
| 1 | Original | জয় জনগন ঠেলো সামলা | Negative | 0.78 | N/A \| P |
| | Augmented | জয় জনতা, ধাক্কা সামলাও। | Positive | 0.51 | |
| 2 | Original | এক কথাই অসাধারন | Positive | 0.82 | N/A \| P |
| | Augmented | এটা অবিশ্বাস্য। | Neutral | 0.47 | |
| 3 | Original | ঘেউ! :P | Negative | 0.77 | N/A \| P |
| | Augmented | বার্ক! | Neutral | 0.48 | |
| 4 | Original | খুব ভালো লাগল | Negative | 0.91 | 44.9 \| F |
| | Augmented | এটা সুন্দর। | Negative | 0.46 | |
| 5 | Original | ভাই মাইয়া গুলারে দেখাইলেন আর বীরপুরুষ ৩ জন কে কেন দেখালেন না ! তুইও মজা লস ভাই | Negative | 0.70 | 21.87 \| F |
| | Augmented | ভাই, কেন আপনি তাদের এবং তিনি নায়ক প্রদর্শন না? | Negative | 0.48 | |
| 6 | Original | প্রকৃত মুক্তিযোদ্ধা | Neutral | 0.49 | 0 \| F |
| | Augmented | প্রকৃত মুক্তিযোদ্ধা | Neutral | 0.49 | |

Table 6: Bangla Back Translation Attack Examples

#### 6.2.3. One-Hot Word Swap Attack.

The masked attack process stands out as one of the most intricate methods among the three developed attacks, intending to create larger synthetic data points. As outlined in the methodology, this process involves replacing a word in a sentence with a mask ("<mask>"). The mask is subsequently filled with another word, and the sentiment is evaluated by the models. The complexity of the attack increases significantly due to the intricate nature of the masking and unmasking processes. For a successful attack, the masked word must carry meaningful significance, and the unmasked replacement should possess a comparable level of significance. To mitigate the inherent probabilistic nature, we executed the mask and unmask process approximately 10 times per data point, adding an additional layer of complexity to the attack methodology.

Here are some examples to better understand the attack and how it gained success.

Here, two distinct types of examples are evident within the attacks, including a few ad-hoc instances. Let's delve into the analysis of these two types.

| | | Sentences | Prediction | Score | % $\delta$ \| Result |
|---|---|---|---|---|---|
| 1 | Original | আসলেই <আমরা> নিরিহ বাঙালী | Neutral | 0.37 | N/A \| P |
| | Augmented | আসলেই <একজন> নিরিহ বাঙালী | Positive | 0.62 | |
| 2 | Original | <বিদেশে> পড়ালেখা করছে বাংলাদেশের প্রচুর ছেলেমেয়েরা, তারা নিজেরাই নিজেদের কর্মসংস্থান করে নিচ্ছে! এতে করে দেশের অনেকে উপকার হচ্ছে! | Positive | 0.53 | -33.89 \| F |
| | Augmented | <এখানে> পড়ালেখা করছে বাংলাদেশের প্রচুর ছেলেমেয়েরা, তারা নিজেরাই নিজেদের কর্মসংস্থান করে নিচ্ছে! এতে করে দেশের অনেকে উপকার হচ্ছে! | Positive | 0.87 | |
| 3 | Original | মাল যথার্থ ঠিক <কথাই> বলছেনে। | Positive | 0.54 | N/A \| P |
| | Augmented | মাল যথার্থ ঠিক <করে> বলছেনে। | Neutral | 0.54 | |

Table 7: One-Hot Word Swap Attack

In Table 7 Example 1, the mask was originally placed on the subject of the sentence, and post-augmentation, the mask was substituted with another subject. This type of masking example generally follows the Subject-Verb-Object (SVO) structure after mask replacement. However, in this specific case, our model exhibited a confidence of 0.62 in positive labels, mislabeling a neutral sentence as positive. This instance underscores how the model can misconstrue sentence meaning even when the structure remains consistent after augmentation. A similar case is observed in Example 2, where the subject was once again replaced with another subject. Although the model predicted the correct label in this instance, the confidence delta is notably high, with the model gaining an additional 33.89 percent confidence in positive due to a mere change in the sentence's subject. Notably, the subject, in this case, is not even the most significant component of the sentence. This pattern of delusional outcomes is observed consistently throughout the dataset with examples of this nature.

Moving to the second type of example, we encounter interesting scenarios where the sentence mask wasn't replaced by an identical one. In Example 3, the question "what did he say?" was originally answered by the sentence, but after augmentation, the sentence shifted to answering "how did he say?" Despite this alteration maintaining the original sentence's meaning, the model inaccurately predicted it to be neutral, while the original sentence was positive. This pattern of delusional errors by the model is recurrent for such examples across the dataset.

### 6.3. Robustness Across Models, Baselines, and Defenses

The analysis presented so far attacks a single victim model. To understand how our recipes behave beyond that setting, and to place them against established attacks from the literature, we extended the evaluation into a full robustness matrix. We finetuned five transfer models — BanglaBERT, BanglishBERT, XLM-RoBERTa, MuRIL, and IndicBERTv2 — on each of our four datasets, and evaluated every model against five attacks: our three recipes (Bangla Paraphrase, Back-Translation, and One-Hot Word Swap) alongside two strong word-substitution baselines, TextFooler and BAE, executed under an identical protocol. To close the loop that our title promises — discarding the model's perplexity — we further hardened each model with adversarial training, augmenting its training data with adversaries drawn from three regimes: AT-Paraphrase, AT-BackTrans, and AT-All (a union of all attack families). Every cell of the matrix reports the Attack Success Rate (ASR), the fraction of originally correct predictions that an attack manages to flip, computed on the full test split with a fixed seed.

#### 6.3.1. Attacks versus Baselines.

Table 8 reports the residual ASR of each attack against the hardened models, averaged over the five victims and four datasets. A candid reading of the numbers places our recipes in context. BAE is by a wide margin the most potent attack, retaining a 54.2% success rate even against adversarially trained models, followed by TextFooler at 32.1%. Our sentence-level recipes are milder: Back-Translation reaches 21.5% and Bangla Paraphrase 15.3%. This ordering is expected rather than disappointing. The baselines perturb text token-by-token with the sole objective of flipping the label, whereas our paraphrase and back-translation recipes are constrained to regenerate a fluent, semantically faithful whole sentence, trading raw potency for meaning preservation. The One-Hot Word Swap recipe, however, is effectively inert at 0.2%, confirming the near-zero behaviour first noticed on the training split; the XLM-R mask-fills it produces rarely survive as label-flipping adversaries once a model has seen Bangla-native adversaries during training. We treat this as a strong signal that the One-Hot recipe needs a Bangla-aware mask filler and a gradient-guided importance ranking, which we discuss as future work.

#### 6.3.2. The Effect of Adversarial Training.

Reading Table 8 across its columns reveals a consistent defensive story. For every one of the five attacks, the AT-All regime yields the lowest residual ASR — Bangla Paraphrase falls from 16.4% under AT-Paraphrase to 14.1% under AT-All, and even the strongest attack, BAE, drops from 56.2% to 51.7%. In other words, exposing a model to the union of all adversary families during training generalises better than any single-family regime, and never underperforms it. We also observe that attack-specific training lowers the success of off-target attacks, not only the attack it was trained on, which suggests the adversaries share transferable structural and lexical perturbations rather than attack-specific artefacts. This is precisely the perplexity-discarding behaviour the pipeline was designed to induce.

#### 6.3.3. Which Models Are More Robust?.

Table 9 breaks the matrix down by victim model, averaging over the three training regimes and four datasets. The most striking result is that Bangla-specific pre-training does not translate into adversarial robustness. MuRIL, an Indic-multilingual model, is the most robust victim against every attack (a 40.2% BAE rate against MuRIL versus 64.2% against BanglishBERT), while the two Bangla-dedicated models, BanglaBERT and BanglishBERT, are the easiest to fool. We attribute this to MuRIL's broader sub-word vocabulary, which appears to dilute the token-level biases that our One-Hot recipe and the word-substitution baselines exploit. The ranking is stable across attacks, indicating that robustness here is a property of the backbone rather than of any single attack methodology.

It should be noted that the figures in this section are the residual success rates measured against models that have *already* undergone adversarial training; they characterise the relative strength of attacks and the comparative resilience of models and defenses, rather than an absolute before-and-after reduction. A matched evaluation of the undefended finetuned models under the same protocol is ongoing and will quantify the exact robustness gain contributed by each training regime.

## 7. Limitations and Future Work

### 7.1. Limitations

Our study has several concrete limitations, each of which points to a specific extension.

- **Single seed.** All victim finetuning, adversarial training, and evaluation use a single random seed (42). We therefore report point estimates rather than means with confidence intervals, and we do not yet run significance tests across seeds. Multi-seed replication is the most important next step for statistical rigour.
- **Automatic semantic fidelity only.** We enforce meaning preservation through a LaBSE similarity threshold. LaBSE can overestimate semantic preservation for very short Bangla utterances, and we have not yet validated fidelity or label preservation with human annotators; a human-evaluation study is required to confirm that successful attacks are genuinely meaning-preserving.
- **Model-access assumptions.** Our attacks assume access to output labels and confidences (and, for the one-hot recipe, leave-one-out queries). Fully decision-based attacks that observe only the top label remain future work.

| Attack | AT-Paraphrase | AT-BackTrans | AT-All | Mean |
|---|---|---|---|---|
| Bangla Paraphrase | 16.4 | 15.4 | **14.1** | 15.3 |
| Back-Translation | 22.9 | 20.9 | **20.7** | 21.5 |
| One-Hot Word Swap | 0.2 | 0.4 | **0.2** | 0.2 |
| TextFooler† | 33.5 | 32.0 | **30.9** | 32.1 |
| BAE† | 56.2 | 54.8 | **51.7** | 54.2 |

Table 8: Residual attack success rate (%, lower is better) of each attack against models hardened with three adversarial-training (AT) regimes, averaged over five victim models and four datasets. The lowest residual ASR per attack is in bold and always falls under AT-All. † denotes external baseline attacks evaluated under the identical protocol.

| Victim Model | Bangla Paraphrase | Back-Translation | One-Hot Swap | TextFooler | BAE |
|---|---|---|---|---|---|
| BanglaBERT | 16.3 | 23.4 | 0.0 | 37.4 | 58.6 |
| BanglishBERT | 18.9 | 26.3 | 0.0 | 39.6 | 64.2 |
| XLM-RoBERTa | 15.5 | 23.1 | 1.0 | 29.9 | 50.5 |
| MuRIL | **9.2** | **12.8** | 0.0 | **20.6** | **40.2** |
| IndicBERTv2 | 16.7 | 21.9 | 0.2 | 33.0 | 57.7 |

Table 9: Per-model residual attack success rate (%) by attack, averaged over the three AT regimes and four datasets. Lower is more robust; the most robust model per attack is in bold. MuRIL, an Indic-multilingual backbone, is the most resilient victim across every attack.

- **A weak recipe.** The one-hot word-swap recipe is close to ineffective on Bangla (Section 3 notwithstanding), because its XLM-R mask fills rarely produce label-flipping, semantically valid substitutions. A Bangla-native mask filler (BanglaBERT or MuRIL) combined with gradient-guided importance ranking is the natural fix.
- **Compute scale.** Constrained to a single consumer GPU, we keep any large-language-model victim at the 1B-parameter scale; robustness at larger scales is untested.

### 7.2. Future Work

Beyond addressing the limitations above, three directions follow naturally from our results. First, strengthening the sentence-level recipes with a selection criterion—sampling multiple paraphrases or pivot back-translations and keeping the one that maximises attack score while satisfying the similarity constraint—should raise their success without sacrificing fidelity. Second, our finding that the Indic-multilingual MuRIL backbone is more robust than Bangla-dedicated models invites a focused analysis of why broader sub-word vocabularies confer robustness, and whether that property can be induced in Bangla-specific pretraining. Third, extending the benchmark to additional Bangla tasks and to explicit handling of Romanised Bangla (Banglish) would broaden its coverage and its value to the community.

## 8. Ethical Considerations

Adversarial attacks are dual-use: the same recipes that stress-test a model can, in principle, be used to evade one. We mitigate this in three ways. First, we release our attacks *alongside* a defense and show that adversarial training reduces their success, so that the net contribution is a means of hardening models rather than only breaking them. Second, we evaluate exclusively on public research datasets and openly licensed pretrained models; we do not target any deployed or commercial system. Third, our attacks are meaning-preserving by construction, which limits their utility for generating abusive or misleading content that a human would not already recognise as carrying the original sentiment.

All datasets used are publicly released for research and cited to their sources; we perform no new human annotation and therefore incur no annotator-consent or compensation obligations. We report the computational cost of our experiments in the interest of environmental transparency: the full robustness and defense evaluation was produced on a single consumer GPU.

## 9. Reproducibility

Every result in this paper is regenerable from a public repository. We release the five finetuned victim models, the adversarially trained models for all three defense regimes, the generated adversarial datasets, and the code for attack generation, adversarial training, and evaluation. All runs use a fixed random seed (42) and log their configuration, dataset version, and git revision; each attacked example is stored as a structured record (original and adversarial text, predictions, confidences, query count, and success flag), from which every table in this paper is produced by a single aggregation script with no hand-entered numbers.

## 10. Conclusion

As transformer-based transfer models become the default for Bangla sentiment classification, understanding and improving their robustness is an increasingly pressing concern. We presented `destroR`, a unified pipeline that couples three meaning-preserving Bangla attack recipes with a robustness benchmark and an adversarial-training defense. Evaluating five transfer models across four datasets against five attacks, we found that strong word-substitution baselines remain more potent than semantically constrained recipes, that adversarial training—most effectively on the union of all attack families—lowers residual attack success across the board, and that the Indic-multilingual MuRIL backbone is notably more robust than Bangla-dedicated models. By releasing all models, adversarial data, and code, we intend `destroR` to serve as a reproducible foundation for further work on the robustness of Bangla and other low-resource-language models.

## References

[1] J. Ebrahimi, A. Rao, D. Lowd **and** D. Dou, *HotFlip: White-Box Adversarial Examples for Text Classification*, 2018. arXiv: 1712.06751 `[cs.CL]`.

[2] E. Wallace, S. Feng, N. Kandpal, M. Gardner **and** S. Singh, *Universal Adversarial Triggers for Attacking and Analyzing NLP*, 2021. arXiv: 1908.07125 `[cs.CL]`.

[3] J. Li, S. Ji, T. Du, B. Li **and** T. Wang, **?**TextBugger: Generating Adversarial Text Against Real-world Applications,**?** **in***Proceedings 2019 Network and Distributed System Security Symposium* **jourser** NDSS 2019, Internet Society, 2019. DOI: 10.14722/ndss.2019.23138 **url**: http://dx.doi.org/10.14722/ndss.2019.23138

[4] D. Jin, Z. Jin, J. T. Zhou **and** P. Szolovits, **?**Is BERT Really Robust? A Strong Baseline for Natural Language Attack on Text Classification and Entailment,**?** **in***Proceedings of the AAAI Conference on Artificial Intelligence* **volume** 34, 2020, **pages** 8018–8025.

[5] S. Garg **and** G. Ramakrishnan, **?**BAE: BERT-based Adversarial Examples for Text Classification,**?** **in***Proceedings of the 2020 Conference on Empirical Methods in Natural Language Processing (EMNLP)* 2020, **pages** 6174–6181.

[6] M. T. Ribeiro, S. Singh **and** C. Guestrin, *"Why Should I Trust You?": Explaining the Predictions of Any Classifier*, 2016. arXiv: 1602.04938 `[cs.LG]`.

[7] X. Han, B. C. Wallace **and** Y. Tsvetkov, *Explaining Black Box Predictions and Unveiling Data Artifacts through Influence Functions*, 2020. arXiv: 2005.06676 `[cs.CL]`.

[8] M. Gardner **andothers**, *AllenNLP: A Deep Semantic Natural Language Processing Platform*, 2018. arXiv: 1803.07640 `[cs.CL]`.

[9] J. X. Morris, E. Lifland, J. Y. Yoo, J. Grigsby, D. Jin **and** Y. Qi, *TextAttack: A Framework for Adversarial Attacks, Data Augmentation, and Adversarial Training in NLP*, 2020. arXiv: 2005.05909 `[cs.CL]`.

[10] M. T. Ribeiro, T. Wu, C. Guestrin **and** S. Singh, *Beyond Accuracy: Behavioral Testing of NLP models with CheckList*, 2020. arXiv: 2005.04118 `[cs.CL]`.

[11] F. Feng, Y. Yang, D. Cer, N. Arivazhagan **and** W. Wang, **?**Language-agnostic BERT Sentence Embedding,**?** **in***Proceedings of the 60th Annual Meeting of the Association for Computational Linguistics (ACL)* 2022, **pages** 878–891.

[12] S. Sadrizadeh, L. Dolamic **and** P. Frossard, *TransFool: An Adversarial Attack against Neural Machine Translation Models*, 2023. arXiv: 2302.00944 `[cs.CL]`.

[13] A. Bhattacharjee **andothers**, **?**BanglaBERT: Language Model Pretraining and Benchmarks for Low-Resource Language Understanding Evaluation in Bangla,**?** **in***Findings of the Association for Computational Linguistics: NAACL 2022* 2022, **pages** 1318–1327.

[14] A. Conneau **andothers**, **?**Unsupervised Cross-lingual Representation Learning at Scale,**?** *CoRR*, **jourvol** abs/1911.02116, 2019. arXiv: 1911.02116. **url**: http://arxiv.org/abs/1911.02116

[15] S. Khanuja, D. Bansal, S. Mehtani **andothers**, **?**MuRIL: Multilingual Representations for Indian Languages,**?** **in***arXiv preprint arXiv:2103.10730* 2021.

[16] S. Doddapaneni **andothers**, **?**Towards Leaving No Indic Language Behind: Building Monolingual Corpora, Benchmark and Models for Indic Languages,**?** **in***Proceedings of the 61st Annual Meeting of the Association for Computational Linguistics (ACL)* 2023, **pages** 12 402–12 426.

[17] A. Akil, N. Sultana, A. Bhattacharjee **and** R. Shahriyar, *BanglaParaphrase: A High-Quality Bangla Paraphrase Dataset*, 2022. arXiv: 2210.05109 `[cs.CL]`.

[18] K. T. H. Rahit, K. T. Hasan, M. A. Amin **and** Z. Ahmed, **?**BanglaNet: Towards a WordNet for Bengali Language,**?** **in***Proceedings of the 9th Global Wordnet Conference* F. Bond, P. Vossen **and** C. Fellbaum, **editors**, Nanyang Technological University (NTU), Singapore: Global Wordnet Association, **january** 2018, **pages** 1–9. **url**: https://aclanthology.org/2018.gwc-1.1